\documentclass{article}

\usepackage[preprint]{neurips_2026}
\usepackage{wrapfig}
\usepackage[utf8]{inputenc} % allow utf-8 input
\usepackage[T1]{fontenc}    % use 8-bit T1 fonts
\usepackage{hyperref}       % hyperlinks
\usepackage{url}            % simple URL typesetting
\usepackage{booktabs}       % professional-quality tables
\usepackage{amsfonts}       % blackboard math symbols
\usepackage{nicefrac}       % compact symbols for 1/2, etc.
\usepackage{microtype}      % microtypography
\usepackage{xcolor}         % colors
\usepackage{graphicx}
\usepackage{algorithm}
\usepackage{algpseudocode}
\usepackage{amsmath}
\usepackage{amssymb}
\newcommand{\pangu}{openPangu}
\newcommand{\modelname}{\pangu~Embedded-1B}
\usepackage{hyperref}
\usepackage[textsize=tiny]{todonotes}
\usepackage[commandnameprefix=always]{changes}
\usepackage{microtype}
\usepackage{graphicx}
\usepackage{subcaption}
\usepackage{multirow}
\usepackage{booktabs} % for professional tables
\usepackage{tabularx,array}
\usepackage{arydshln}
\usepackage{romannum}
\title{Near-Policy: Accelerating On-Policy Distillation via Asynchronous Generation and Selective Packing}

\author{%
  Miao Rang\thanks{Equal contribution.} $^1$ \quad
  Zhenni Bi\footnotemark[1]  $^1$  \quad
  Hang Zhou$^2$ \quad
  Kai Han$^1$  \quad  Xuechun Wang$^1$ \\
  \textbf{An Xiao$^1$ \quad Xinghao Chen$^1$ \quad Yunhe Wang$^1$ \quad
  Hanting Chen$^1$\thanks{Corresponding author. Email: \texttt{chenhanting@huawei.com}}}
  \\
  $^1$Huawei Technologies, $^2$Tianjin University
}

\begin{document}
\pagenumbering{arabic}

\maketitle

\begin{abstract}
  Standard knowledge distillation for autoregressive models often suffers from distribution mismatch. While on-policy methods mitigate this by leveraging student-generated outputs, they rely on computationally expensive Reinforcement Learning (RL) frameworks. To improve efficiency, we propose \textbf{Near-Policy Distillation (NPD)}, an asynchronous approach that decouples student generation from training. This reformulation enables Supervised Fine-Tuning (SFT) with sequence packing. However, asynchronous updates inevitably introduce policy lag and sample noise, which can cause the behavior to drift from near-policy toward off-policy. To counteract this without sacrificing efficiency, NPD integrates sparse student updates and the \textbf{$\Delta$-IFD} filtering mechanism, a heuristic sample selection mechanism that empirically stabilizes the optimization trajectory. By filtering extreme out-of-distribution samples, $\Delta$-IFD prevents noise from dominating the gradients, ensuring updates remain within a safe proximal learning zone. Empirically, the NPD framework achieves a \textbf{8.1}${\times}$ speedup over on-policy baselines and outperforms SFT by \textbf{8.09$\%$}. Crucially, by effectively narrowing the exploration space for subsequent RL, our method enables openPangu-Embedded-1B to reach a state-of-the-art score of \textbf{68.73}$\%$, outperforming the substantially larger Qwen3-1.7B. Codes will be released soon.
\end{abstract}

\section{Introduction}
% Crucially, by effectively narrowing the exploration space, the method allows openPangu-Embedded-1B to reach a state-of-the-art score of \textbf{68.73}$\%$ when combined with RL

Large Language Models (LLMs) have revolutionized artificial intelligence by leveraging self-attention mechanisms and massive-scale pre-training to capture hierarchical linguistic patterns, semantic relationships, and cross-domain knowledge~\cite{achiam2023gpt}. Open-sourced models such as LLaMA~\cite{touvron2023llama}, DeepSeek~\cite{liu2024deepseek}, Qwen~\cite{bai2023qwen,yang2025qwen3}, and openPangu~\cite{chen2025pangu,tang2025pangu} have demonstrated exceptional capabilities in complex tasks, including reasoning and multilingual understanding. However, their success relies heavily on colossal parameter counts (e.g., DeepSeek-V3~\cite{liu2024deepseek} possesses 671B parameters) and extensive computational resources. The prohibitive energy consumption and hardware requirements of these models present critical barriers to real-world deployment, particularly in scenarios demanding on-device processing, privacy preservation, or low-latency responsiveness.

Standard Knowledge Distillation (KD) approaches~\cite{hinton2015distilling,sanh2019distilbert} typically utilize Kullback-Leibler Divergence (KLD) loss to align the student’s output distribution with the teacher’s on a fixed dataset. While this inherits the architectural efficiency of Supervised Fine-Tuning (SFT) and improves performance over purely supervised methods, it suffers from a fundamental \textit{distribution mismatch}: the student is trained on teacher-forced sequences but generates autoregressive sequences during inference. To mitigate this, on-policy methods~\cite{agarwal2024policy} train the student on its own Student-Generated Outputs (SGOs). Although effective in correcting distribution shift, these methods rely on synchronous RL-like frameworks that necessitate simultaneous teacher-student inference. Consequently, they inherit the inefficiencies of RL, specifically the inability to utilize sequence packing and the low token utilization rate caused by continuous generation.

To overcome these limitations, we propose Near-Policy Distillation. This asynchronous framework fundamentally decouples SGOs generation from the gradient update process. By pre-computing SGOs and teacher logits, NPD approximates the distribution-matching benefits of on-policy distillation while reformulating the training phase as SFT. Crucially, this allows us to leverage sequence packing, thereby maximizing training throughput. However, the asynchronous nature of NPD introduces a new challenge: \textit{policy lag}. Since the data generation policy differs from the current training policy, samples may become noisy or drift toward off-policy distributions, potentially destabilizing the student's alignment.

To counteract this drift without sacrificing efficiency, we employ a dual strategy. First, we implement sparse student updates, ensuring that the generation policy never deviates significantly from the training policy. Second, we introduce a dynamic filtering mechanism based on \textbf{$\Delta$-IFD}. By leveraging the Instruction-Following Difficulty (IFD)~\cite{li2024quantity} metric, we estimate the discrepancy between the teacher and the evolving student policy. This mechanism acts as a gatekeeper, empirically stabilizing the optimization trajectory by filtering out extreme out-of-distribution samples. This prevents high-variance noise from dominating the gradients and ensures the student updates within a safe proximal learning zone, maintaining a near-policy state despite the asynchronous updates. Empirically, the NPD framework yields an \textbf{8.09\%} performance improvement across the distillation phase.

Finally, we investigate the synergy between distillation and reinforcement learning. We demonstrate that models trained via NPD serve as superior initialization points for RL. By effectively constraining the exploration space, NPD enables the subsequent RL stage to achieve significantly higher precision. 

% Empirically, our full-stack framework achieves a \textbf{7.9$\times$} speedup over on-policy baselines.

Our main contributions are summarized as follows:

\begin{itemize}

    \item We propose \textbf{Near-Policy Distillation}, an asynchronous framework that decouples generation from training to enable sequence packing, achieving the performance benefits of on-policy distillation with the efficiency of off-policy training\textbf{(8.1$\times$ speedup)}.

    \item We address the policy lag inherent in asynchronous updates by employing sparse student updates alongside the \textbf{$\Delta$-IFD} filtering mechanism. By filtering extreme out-of-distribution samples, it prevents noise-dominated gradients and keeps updates within a safe proximal learning zone, thereby empirically stabilizing the optimization trajectory.

    \item We demonstrate that NPD unlocks the full potential of subsequent RL by narrowing the search space. This powerful combination enables our \pangu~Embedded-1B model to achieve a state-of-the-art score of \textbf{68.73\%}, surpassing the substantially larger Qwen3-1.7B (63.69\%).

\end{itemize}

\section{Related Works}

Knowledge Distillation (KD)~\cite{Bucila2006ModelC,hinton2015distilling} transfers capabilities from a high-capacity teacher to a compact student. In the context of autoregressive LLMs, methodologies primarily diverge into RL-based (sample-wise distillation) and SFT-based (sequence-packed distillation) paradigms.
In RL-style (sample-wise) distillation, the student is optimized under its \emph{own} trajectory distribution to reduce the training--inference mismatch that arises when learning from teacher-generated sequences but deploying on student-generated ones. MiniLLM~\cite{gu2023minillm} argues that forward-KL distillation (maximum likelihood under the teacher distribution) may over-cover low-probability regions and harm generation quality, and therefore adopts a reverse-KL objective with an effective on-policy training procedure~\cite{gu2023minillm}. GKD~\cite{agarwal2024onpolicydistillationlanguagemodels} makes this principle explicit by sampling student trajectories and distilling teacher token-level guidance on them, ensuring updates target the inference-time state distribution~\cite{agarwal2024onpolicydistillationlanguagemodels}. Building on this line, DistiLLM and DistiLLM-2~\cite{ko2024distillmstreamlineddistillationlarge,ko2025distillm2contrastiveapproachboosts} improve objective design (e.g., skew-KL and contrastive formulations) and data efficiency via adaptive reuse of student-generated samples, reducing the cost of online sampling and teacher inference~\cite{ko2024distillmstreamlineddistillationlarge,ko2025distillm2contrastiveapproachboosts}. Overall, on-policy distillation better corrects mismatch under the student's distribution, but its throughput is often limited by online generation and teacher evaluation overhead.

% \textbf{SFT-based Distillation}, conversely, resembles classical offline learning. Methods such as standard KD~\cite{hinton2015distilling} and Sequence-level KD (SeqKD)~\cite{kim2016sequencelevelknowledgedistillation} train the student to mimic teacher logits or synthetic sequences. These approaches are highly scalable and integrate naturally with \textit{sequence packing}, offering substantial throughput advantages. However, as they rely predominantly on teacher-forced trajectories, they often fail to mitigate distribution mismatch, leaving the student vulnerable to error propagation during inference.
% \textbf{SFT-based (sequence-packed) distillation}, in contrast, follows a classical offline learning setup.
% Methods such as standard KD~\cite{hinton2015distilling} and sequence-level KD (SeqKD)~\cite{kim2016sequencelevelknowledgedistillation} train the student to match teacher logits on a fixed corpus and/or to learn from teacher-generated synthetic sequences.
% Because the objective is purely supervised on pre-collected trajectories, these approaches are highly scalable and are \emph{amenable} to throughput optimizations such as sequence packing.
% However, since training typically conditions on reference trajectories (ground-truth or teacher-generated) rather than student rollouts, offline distillation does not explicitly correct the state-distribution shift at inference time and may therefore still suffer from exposure bias and error accumulation during generation.
SFT-based (sequence-packed) distillation follows a classical offline learning setup. Methods such as standard KD~\cite{hinton2015distilling} and SeqKD~\cite{kim2016sequencelevelknowledgedistillation} train the student to match teacher logits and/or learn from teacher-generated sequences on a fixed corpus. This purely supervised formulation is highly scalable and naturally compatible with throughput optimizations such as sequence packing. However, because training conditions on reference trajectories rather than student rollouts, it does not explicitly correct inference-time state-distribution shift and may still suffer from exposure bias and error accumulation. To bridge this gap, we propose Near-Policy Distillation. By decoupling generation from optimization via an asynchronous pipeline, NPD distills teacher guidance on student-generated trajectories while retaining SFT-style sequence packing for high throughput, improving efficiency and maintaining alignment with the evolving student policy.

\begin{figure*}[t]
    \centering
    % width 可以设置为具体的厘米数 (e.g., 5cm) 或页面宽度的比例 (e.g., 0.8\textwidth)
    \includegraphics[width=0.8\textwidth]{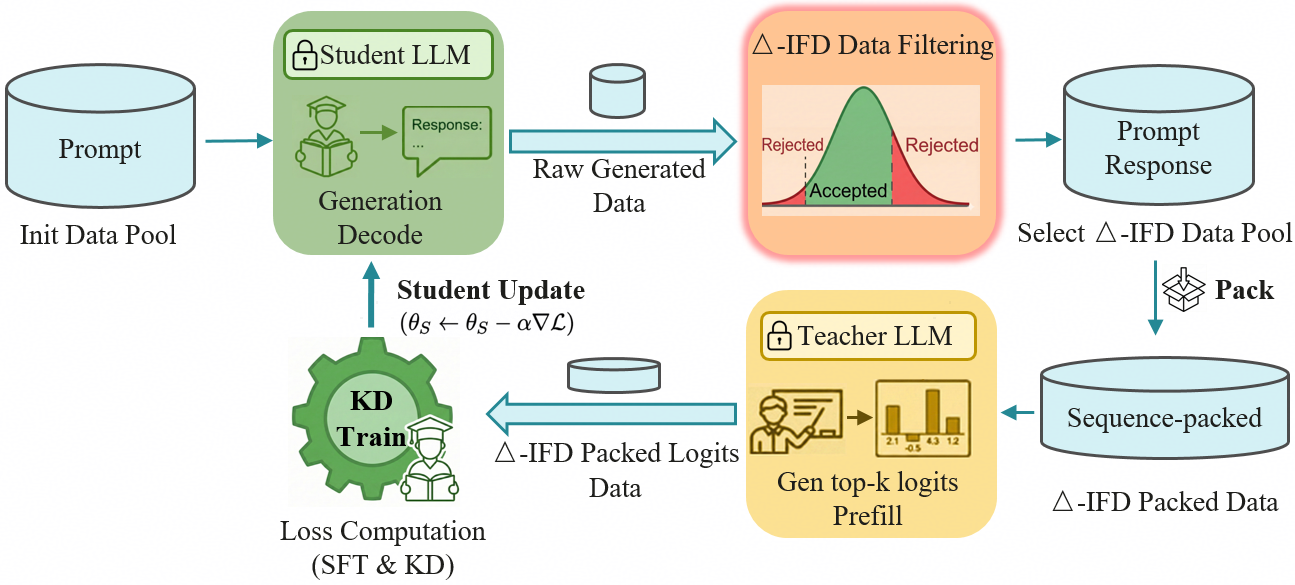}
    \caption{\textbf{Overview of the Near-Policy Distillation Loop with $\Delta$-IFD data selection and student updates.} The pipeline consists of three phases: (1) \textbf{Generation \& Filtering:} The Student LLM generates raw responses from the initial prompt pool, which are then filtered using the $\Delta$-IFD metric. Samples falling outside the $\Delta_{\text{IFD}}$ threshold are rejected to ensure data quality. (2) \textbf{Teacher Annotation:} The selected samples are sequence-packed, and the Teacher LLM performs a prefill step to generate top-$k$ logits. (3) \textbf{Student Updates:} The Student LLM is updated using a combined loss of SFT and KD derived from the teacher's logits.}
    \label{fig:my_image_label}
    \vspace{-0.4cm}
\end{figure*}
\vspace{-0.4cm}

\section{Method}
% In this section, we present the technical details of our proposed Near-Policy Knowledge Distillation (NPD) framework, as shown in Fig ~\ref{fig:my_image_label}. We first provide a formal definition of the asynchronous distillation paradigm, illustrating how it decouples generation from training to maximize computational throughput. Subsequently, we introduce the $\Delta$-IFD filtering mechanism, designed to curate high-value samples by leveraging teacher-student discrepancy. Finally, we discuss the optimization objective and the synergistic integration of NPD with Reinforcement Learning.

In this section, we detail the NPD framework (Fig~\ref{fig:my_image_label}). We first define the asynchronous paradigm, which decouples generation from training to maximize throughput. Subsequently, we introduce the $\Delta$-IFD mechanism designed to curate high-value samples by leveraging teacher-student discrepancy. Finally, we discuss the optimization objective and the synergistic integration of NPD with Reinforcement Learning.

\subsection{Near-Policy Knowledge Distillation}

% We propose \textbf{Near-Policy Distillation (NPD)}. 
As outlined in Algorithm~\ref{alg:iterative_kd}, Our core strategy involves \textit{decoupling} the student's generation from the gradient update loop. This design bridges the distribution mismatch—a key advantage of on-policy methods—while unlocking the high training throughput typical of SFT.

The workflow proceeds in three streamlined stages: (1) \textbf{Generation:} We utilize high-throughput batch inference to generate student responses; (2) \textbf{Gen Logits:} These sequences are packed and fed into the teacher model, which computes logits via a highly efficient parallel prefill process; (3) \textbf{Optimization:} Finally, distillation is performed within a standard SFT architecture utilizing an auxiliary KD loss. 

% Specifically, the workflow proceeds as follows: First, we utilize high-throughput batch inference to efficiently generate student responses. These sequences are then packed and fed into the teacher model, which computes logits via a highly efficient parallel prefill process (avoiding autoregressive decoding). Finally, distillation is performed within a standard SFT architecture, requiring only an auxiliary KD loss term. This streamlined pipeline ensures operational simplicity and scalability. Below, we detail the two critical components of the data preparation phase.

\textbf{Student-Driven Response Generation.} 
To align the training distribution with the student's evolving policy, we employ the current student model (parameterized by $\theta_{S}$) to generate responses for the prompts dataset $\mathcal{D}_{x}$. We leverage high-performance inference engines (e.g., vLLM) to perform batch inference, producing a set of student-generated trajectories $\mathcal{D}_{S} = \{(x, y) \mid y \sim p_{\theta}(\cdot|x)\}$. This step ensures that the subsequent teacher guidance is grounded directly in the student's current exploration space, effectively mitigating the distribution shift often observed in standard KD.

\textbf{Teacher Logits via Packed Parallel Prefill.} 
Unlike on-policy methods that require synchronized teacher generation, our decoupled framework allows the teacher to act as a pure evaluator. Crucially, this enables the application of sequence packing during the annotation phase. We concatenate multiple student responses into fixed-length sequences, allowing the teacher to process them via parallel forward pass (prefill) rather than costly autoregressive decoding. For each position $t$ in a sequence $y$, the teacher computes the probability distribution conditioned on the prefix $y_{<t}$. To balance storage efficiency with distillation quality, we record only the top-$k$ logits rather than the full vocabulary distribution. This strategy preserves the teacher's distributional uncertainty within the student's relevant search space, providing rich, fine-grained supervision while maximizing computational resource utilization.

\textbf{KD Train with a Composite Loss.} In the training phase, the student model is updated by minimizing a composite loss function that jointly incorporates standard supervised learning and knowledge distillation. The total objective $\mathcal{L}_{\text{total}}$ is formulated as a weighted combination of the Cross-Entropy (CE) loss and KD loss:

% In the KD training phase, the student model is optimized using a composite loss function that synergistically combines standard supervised learning with knowledge distillation. The total loss $L_{total}$ is formulated as the weighted sum of the Cross-Entropy (CE) loss $L_{CE}$ and the knowledge distillation loss $L_{KD}$:

\begin{equation}
L_{total} = (1 - \lambda) \cdot L_{CE} + \lambda \cdot L_{KD},
\end{equation}
\vspace{-0.6cm}

where $\lambda \in [0, 1]$ is a hyperparameter regulating the trade-off between the hard-label supervision ($\mathcal{L}_{\text{CE}}$) and the teacher's soft distributional guidance.

% The core of the distillation process is the minimization of the KL divergence, denoted $D_{KL}(P \parallel Q)$, is an asymmetric measure that quantifies how a probability distribution $Q$ (from the student) differs from a reference probability distribution $P$ (from the teacher).
% For discrete distributions over a vocabulary of classes C, it is defined as:

% \begin{equation}
% D_{KL}(P \parallel Q) = \sum_{c \in C} P(c) \log \frac{P(c)}{Q(c)},
% \end{equation}
% By minimizing the KL divergence, the student's output distribution ($Q$) is trained to approximate the soft-target distribution ($P$) provided by the teacher. A lower divergence value signifies that the student has successfully learned to mimic the teacher's predictive patterns, effectively internalizing the knowledge transferred during distillation.

To focus the distillation on the most plausible tokens and filter out long-tail noise, we adopt a Top-$k$ distillation objective. Let $\mathcal{V}_{\text{top-}k} \subset \mathcal{V}$ denote the subset of indices corresponding to the $k$ largest logits of the teacher model. The distillation loss is reformulated as the KL divergence computed exclusively over this high-confidence support:

\begin{equation}
\mathcal{L}_{\text{KD}} = \sum_{c \in \mathcal{V}_{\text{top-}k}} p_{\text{T}}(c) \log \frac{p_{\text{T}}(c)}{p_{\text{S}}(c)},
\end{equation}

Minimizing this objective aligns the student's predictive distribution with the teacher's dominant predictions, encouraging the student to capture key semantic information while ignoring negligible probability mass in the tail.

% \begin{equation}
% The distillation term, $\mathcal{L}_{\text{KD}}$, minimizes the Kullback-Leibler (KL) divergence between the teacher's and student's distributions. Let $p_{\text{T}} = \text{Softmax}(O_{\text{T}})$ and $p_{\text{S}} = \text{Softmax}(O_{\text{S}})$ denote the probability distributions over the vocabulary $\mathcal{V}$. The loss is defined as:
% $\mathcal{L}_{\text{KD}} = \tau^2 \sum_{c \in \mathcal{V}} \sigma(O_T/\tau)_c \log \frac{\sigma(O_T/\tau)_c}{\sigma(O_S/\tau)_c}$,

% \end{equation}

% where $\lambda$ is a scalar hyperparameter that is a weighting coefficient. It meticulously balances the influence of the direct supervised objective ($L_{CE}$) and the teacher's distributional guidance ($L_{KD}$).

% The distillation term, $\mathcal{L}_{\text{KD}}$, minimizes the Kullback-Leibler (KL) divergence between the teacher's and student's softened probability distributions. Given the teacher logits $O_T$ and student logits $O_S$ for a sequence, this is defined as:
% where $\sigma(\cdot)$ denotes the softmax function, $\tau$ is the temperature scaling factor, and $\mathcal{V}$ represents the vocabulary. Minimizing this divergence encourages the student to emulate the teacher's full distributional uncertainty ($O_T$), thereby capturing fine-grained semantic relationships (dark knowledge) that hard labels alone cannot convey.

% \vspace{-0.6cm}
\begin{wrapfigure}{r}{0.5\textwidth} 
    \vspace{-0.6cm}
    \centering
    % 注意：这里的 \linewidth 是指 wrapfigure 区块内部的宽度。
    % 因为外层已经限制了占据一半版面，所以这里设置为 1\linewidth（或直接 \linewidth）即可填满这一半。
    \includegraphics[width=\linewidth]{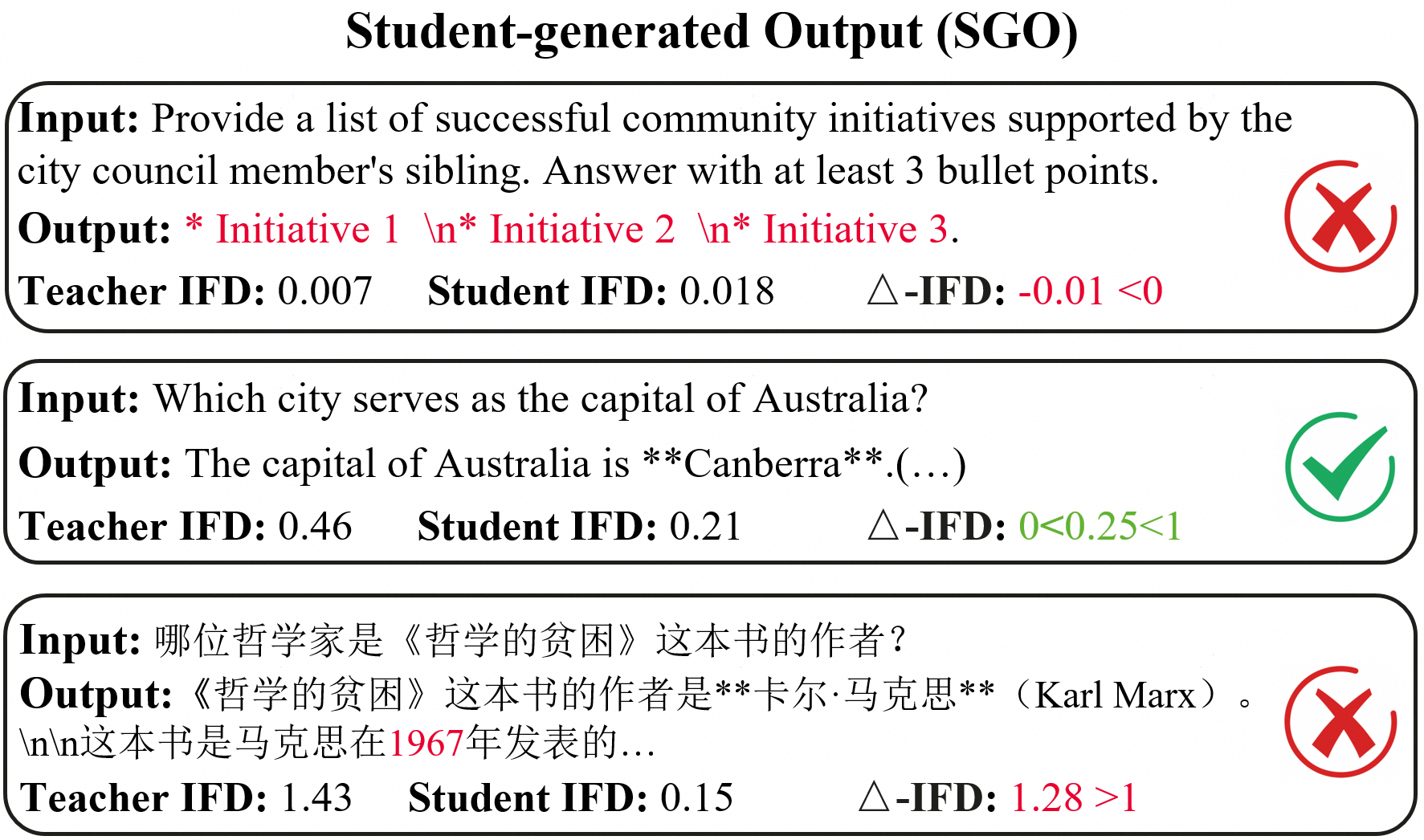}
    \caption{Illustration of Data Curation Zones defined by $\Delta$-IFD.}
    \label{fig:case_analyse}
    \vspace{-0.8cm} % 如果排版过紧或过松，可以微调这个值
\end{wrapfigure}

% \begin{figure}[t] % [t] 表示强制置顶，跨栏图片通常只能放在页顶
%     \centering
%     % width=\textwidth 表示占满整个页面的书写宽度
%     \includegraphics[width=0.5\linewidth]{fig/case.PNG}
%     \caption{Illustration of Data Curation Zones defined by $\Delta$-IFD. }
%     \label{fig:case_analyse}
%     \vspace{-0.6cm}
% \end{figure}

\subsection{Adaptive Data Selection via $\Delta$-IFD}
\label{sec:data_filtering}

In the near-policy setting, response data is generated by the student model itself. While these student-generated outputs align closely with the student's intrinsic distribution, they inevitably introduce quality noise compared to ground-truth data. To empirically filter out noise while retaining training signals that drive improvement, we introduce a heuristic sample selection based on the IFD~\cite{li2024quantity, zhou2024star} metric.

\paragraph{IFD Definition.} 
IFD quantifies the semantic alignment between an instruction $Q$ and a response $A$ by contrasting the unconditioned loss against the conditioned loss. Formally, let $\mathcal{L}_{\theta}(\cdot)$ denote the cross-entropy loss of the sequence. The IFD score is defined as:
\begin{equation}
    \text{IFD}(Q, A) = \frac{\mathcal{L}_{\theta}(A \mid Q)}{\mathcal{L}_{\theta}(A)},
\end{equation}
Intuitively, a lower IFD score indicates higher model confidence and better alignment, whereas a score $\text{IFD} > 1$ implies that the instruction fails to facilitate---or even hinders---the prediction of $A$.

\paragraph{Differential Difficulty ($\Delta$-IFD).}
To capture the \textit{relative learnability} of a sample, we propose the differential IFD metric, denoted as $\Delta_{\text{IFD}}$. This metric measures the cognitive gap between the teacher's evaluation and the student's confidence:
\begin{equation}
    \Delta_{\text{IFD}} = \text{IFD}_{\text{Teacher}} - \text{IFD}_{\text{Student}},
\end{equation}
where $\text{IFD}_{\text{Student}}$ represents the student's intrinsic uncertainty (lower implies higher confidence), and $\text{IFD}_{\text{Teacher}}$ reflects the teacher's acceptability of the reasoning path. Based on $\Delta_{\text{IFD}}$ and a difficulty threshold $\tau$, we categorize student-generated samples into three distinct zones, as shown in Fig~\ref{fig:case_analyse}:
\begin{itemize}
    \item \textbf{The Degenerate Zone ($\Delta_{\text{IFD}} < 0$):} Although student-generated data typically aligns better with the student's own distribution (i.e., $\text{IFD}_{\text{Student}} < \text{IFD}_{\text{Teacher}}$), a negative differential ($\text{IFD}_{\text{Teacher}} < \text{IFD}_{\text{Student}}$) presents an anomaly where the teacher exhibits higher confidence than the student. As illustrated in the first row of Figure~\ref{fig:case_analyse}, this counter-intuitive phenomenon typically arises from \textit{short, meaningless responses} (e.g., trivial patterns or empty sequences). Since the teacher has seen such frequent patterns during pre-training, it assigns them low loss, whereas the student generates them with high uncertainty due to policy degradation. We classify these as degenerate noise and exclude them.
    \item \textbf{The Cognitive Disconnect Zone ($\Delta_{\text{IFD}} > \tau$):} 
    A differential exceeding the threshold $\tau$ indicates that the \textit{teacher model exhibits extreme uncertainty} regarding the response (i.e., $\text{IFD}_{\text{Teacher}} \gg \text{IFD}_{\text{Student}}$). This highlights a severe cognitive discrepancy: while the student generates the content with high confidence, the teacher evaluates it as highly improbable or erroneous. Including such conflicting data creates a substantial mismatch that exerts a detrimental impact on the training process, forcing high-variance optimization steps and destabilizing the training trajectory.
    % potentially confusing the student model's optimization direction.
    \item \textbf{The Proximal Learning Zone ($0 \le \Delta_{\text{IFD}} \le \tau$):} 
    We target samples where the teacher confirms base quality ($\text{IFD}_{\text{Teacher}} \le 1$) and the difficulty gap is moderate. These samples represent ``low-hanging fruit"---challenges that are slightly above the student's current competence but strictly within reach. Distilling logits from this zone serves as a stable learning signal, ensuring the student model updates safely within its current capacity without being overwhelmed by noise.

    % Distilling logits from this zone serves as a beneficial learning signal, helping guide the student in adjusting its decision boundaries.
    % Drawing on the theory of the \textit{Zone of Proximal Development} (ZPD), 
\end{itemize}
Consequently, our final training set is constructed by retaining only samples located within the proximal learning zone. Crucially, this heuristic filtering serves as a robust stabilizer against the policy lag inherent in asynchronous frameworks. Even if the data-generating policy drifts from the current training state, the $\Delta$-IFD criterion effectively intercepts extreme out-of-distribution samples. Empirically (Fig~\ref{fig:policy_latency}), without $\Delta$-IFD, noisy samples trigger high-variance updates, driving the KL divergence to 0.12. In contrast, $\Delta$-IFD effectively intercepts out-of-distribution noise, suppressing the divergence within 0.1. This prevents noise-dominated gradients and stabilizes the optimization trajectory, ensuring updates safely remain within the proximal learning zone.

% Empirically, as shown in Fig ~\ref{fig:policy_latency}, without this filter (Near-Policy w/o $\Delta$-IFD), we observe the KL divergence continuously climbing to 0.12, indicating that noisy or overly difficult samples force high-variance optimization steps. By employing $\Delta$-IFD, the KL divergence is successfully suppressed within 0.01. By preventing the gradient from being dominated by noise, this mechanism empirically stabilizes the optimization trajectory and ensures the student updates within a safe proximal learning zone, regardless of generation latency.

% \textbf{{Multi-Path Reasoning Expansion.} 

% To further push the performance boundaries, we introduce a multi-path reasoning strategy. Under conditions where computational budgets permit, we extend the student's inference process from generating a single greedy path to sampling $K$ independent trajectories for each instruction.

% This expansion significantly broadens the \textbf{sample diversity}, allowing the model to explore a wider region of its potential solution space. Crucially, this strategy synergizes effectively with $\Delta$-IFD filtering mechanism. By creating a diverse pool of candidate responses and rigorously applying the $\Delta_{\text{IFD}}$ criterion, we implement a robust ``\textit{generate-then-filter}'' paradigm. This ensures that the distillation process benefits from a richer variety of reasoning patterns while strictly excluding low-quality noise, ultimately yielding further improvements in distillation precision.

\subsection{Constraining the RL Search Space} 
Reinforcement Learning inherently struggles with expansive search spaces, often resulting in slow convergence and instability during the cold-start exploration phase. While standard protocols typically initiate RL from a SFT checkpoint to mitigate this, we demonstrate that the NPD framework yields a significantly superior initialization.

Our analysis reveals that the distilled model functions as more than a mere weight initialization; it fundamentally \textit{reshapes} the initial policy distribution. By training exclusively on high-quality data filtered via $\Delta_{\text{IFD}}$, the student's probability mass becomes highly concentrated on a manifold of logically correct and instruction-aligned trajectories prior to RL. This effective \textit{pre-conditioning} drastically narrows the exploration space, enabling the subsequent RL stage to bypass inefficient stochastic exploration and focus immediately on fine-grained optimization, thereby unlocking superior asymptotic performance.

\section{Experiments}
\subsection{Experimental Setups}
% \paragraph{Training} In the distillation phase, we reuse the SFT dataset and train the model for 10 epochs. We employ the AdamW optimizer with a weight decay of 0.1 and a gradient clipping threshold of 1.0. To maximize computational efficiency, we employ sample packing with a maximum sequence length of 32,768 tokens. The global batch size is set to 2 million tokens, and the learning rate follows a cosine decay schedule ranging from $1\times 10^{-5}$ to $1\times 10^{-6}$.  openPangu-Embedded-7B~\cite{chen2025pangu} as the teacher and openPangu-Embedded-1B as the student.

\textbf{Knowledge Distillation Setup.} We employ openPangu-Embedded-7B~\cite{chen2025pangu} as the teacher model to distill knowledge into the openPangu-Embedded-1B student model. The training is conducted on the SFT dataset for 10 epochs. We utilize the AdamW optimizer with a weight decay of $0.1$ and a gradient clipping threshold of $1.0$. To maximize computational efficiency, we implement sample packing with a maximum sequence length of $32,768$ tokens. The global batch size is set to 2 \text{M} tokens, and the learning rate follows a cosine decay schedule ranging from $1\times 10^{-5}$ to $1\times 10^{-6}$. Unless otherwise stated, all experiments follow these default configurations.

\textbf{Evaluation.} We evaluate our models on a diverse suite of 11 benchmarks categorized into four primary domains:
(1) \textbf{General Capabilities} are assessed via MMLU, CMMLU~\cite{Li2023cmmlu}, C-Eval~\cite{Huang2023CEvalAM}, IF-Eval~\cite{zhou2023instruction}, and CLUEWSC~\cite{xu2020clue}.
(2) \textbf{Mathematics} is evaluated using GSM8K and MATH-500~\cite{hendrycks2021measuring}.
(3) \textbf{Reasoning} capabilities are tested on DROP~\cite{Dua2019DROPAR} and GPQA-Diamond~\cite{rein2024gpqa}.
(4) \textbf{Code Generation} is measured using MBPP~\cite{austin2021program} and HumanEval~\cite{Chen2021EvaluatingLL}.

\subsection{Comparison with Other Distillation Methods}
To ensure a rigorous and comprehensive evaluation, we benchmark NPD against two distinct distillation paradigms: the efficiency-oriented Sequence-Packed Distillation and the performance-oriented Sample-Wise Distillation.

% \vspace{-0.6cm}
\begin{table}
    \centering
    \caption{Comparison of accuracy and training time between NPD and other distillation methods. The experiments are conducted by distilling the teacher model \textbf{Qwen2.5-Math-7B-Inst ($\mathcal{M}_T$)} into the student model \textbf{Qwen2.5-Math-1.5B ($\mathcal{M}_S$)}. Train Time represents the NPU hours consumed per epoch. Results annotated with $^{\dagger}$ are taken from DistiLLM-2.}
    \label{tab:efficiency_comparison}
    % \vspace{-0.2cm}
    \footnotesize
    \renewcommand{\arraystretch}{1.2} % 增加行高，提升阅读舒适度
    \begin{tabular}{lccc}
        \toprule
        \textbf{Method} & \textbf{GSM8K} & \textbf{MATH500} & \textbf{Train Time} \\
        % \textbf{Train Time (GPU-Hours)} \\
        \midrule
        $\mathcal{M}_T$ (Teacher) & 89.31$^{\dagger}$ & 80.00 & -- \\
        $\mathcal{M}_S$ (Student) & 77.33$^{\dagger}$ & 62.00 & -- \\
        \hdashline[2pt/2pt] % 虚线分割基线与对比方法
         % SFT & 80.06 &  70.80 & \\
         Supervised KD & 78.62 & 75.00  & 1.17 \\
         SeqKD & 82.41 & 72.60 & 2.06 \\
           
        GKD & 80.17 &  73.60 &6.54 \\
        DistiLLM & 81.15 & 73.80 & 12.00 \\
        DistiLLM-2 & 81.27 & 74.20 & 12.00 \\
        \hline
        \textbf{NPD} & \textbf{82.87} & \textbf{76.60} & 1.48 \\
        \bottomrule
    \end{tabular}
  % \vspace{-0.6cm}  
\end{table}

\begin{wraptable}{r}{0.45\textwidth}
    \vspace{-0.5cm} % 消除上方的多余留白
    \centering
    \small
    \caption{Comparison of NPD and sequence-packed distillation methods on a large-scale real-world industrial dataset. AVG denotes the average score across 11 evaluation benchmarks.} % 环绕模式下标题必须极简
    \label{npd_seq}
    \begin{tabular}{lc} 
        \toprule
        \textbf{Method} & \textbf{AVG} \\
        \midrule
        SFT & 56.28 \\
        Supervised KD~\cite{sanh2019distilbert} & 58.79 \\
        SeqKD~\cite{kim2016sequence} & 59.55 \\
        \hline
        \textbf{NPD} & \textbf{64.37} \\ 
        \bottomrule
    \end{tabular}
    \vspace{-0.4cm} % 消除下方的多余留白
\end{wraptable}

\textbf{Comparison with Sample-Wise Distillation.} Due to the prohibitive computational costs associated with sample-wise baselines, we conduct controlled experiments on a 50k-instance subset of MetaQA~\cite{yu2023metamath}. We employ Qwen2.5-Math-7B~\cite{yang2024qwen2} as the teacher and Qwen2.5-Math-1.5B as the student, training all methods for one epoch to ensure strict experimental fairness. As detailed in Table~\ref{tab:efficiency_comparison}, our method demonstrates superiority in both performance and speed. On the GSM8K and Math500 benchmarks, NPD not only surpasses the strong baseline DistiLLM-2~\cite{ko2025distillm2contrastiveapproachboosts} in accuracy but also delivers a substantial efficiency gain. Specifically, under identical training conditions, NPD achieves a $8.1\times$ training speedup compared to DistiLLM-2, validating its capability to deliver on-policy-level performance with off-policy-level efficiency.

\textbf{Comparison with Sequence-Packed Distillation.} We further evaluate the scalability of NPD in a large-scale setting using an internal SFT dataset comprising 1.2 million high-quality samples.

For this experiment, we utilize openPangu-Embedded-7B~\cite{chen2025pangu} as the teacher and openPangu-Embedded-1B as the student. As shown in Table~\ref{npd_seq}, NPD demonstrates superior scalability compared to standard approaches. Crucially, the performance advantage of NPD is significantly amplified when applied to this larger-scale, real-world dataset. It yields substantial gains, outperforming the SFT baseline by 8.09\% and the standard SeqKD approach by 4.83\%. These results, further detailed in Table~\ref{tab:distillation_methold_detail}, confirm that NPD effectively bridges the gap between the efficiency of sequence packing and the distribution-matching benefits of interactive distillation.

% Leveraging the high throughput inherent to our framework, we further evaluate the scalability of NPD on a large-scale proprietary SFT dataset comprising 1.2 million high-quality samples. 
%Capitalizing on the computational efficiency of sequence-packed training, we extend our experiments to a large-scale proprietary SFT dataset comprising 1.2 million high-quality samples, utilizing openPangu-Embedded-7B~\cite{chen2025pangu} as the teacher and openPangu-Embedded-1B as the student. As shown in Table~\ref{tab:performance_avg}, NPD method yields the most significant gains, outperforming the SFT baseline by 5.35$\%$ and the SeqKD approach by 2.1$\%$. Detailed results are provided in Table~\ref{tab:distillation_methold_detail}.

\vspace{-0.2cm}
\subsection{Efficiency and Performance with NPD}

To rigorously evaluate the advantages of NPD, we conduct a comparative study against standard on-policy distillation on the same dataset. We assess performance under two distinct constraints:  (1) \textbf{sample efficiency} (fixed training steps) and (2) \textbf{computational efficiency} (fixed wall-clock time).

\begin{figure*}[t] % [t] 表示强制置顶，跨栏图片通常只能放在页顶
    \centering
    % width=\textwidth 表示占满整个页面的书写宽度
    \includegraphics[width=0.8\linewidth]{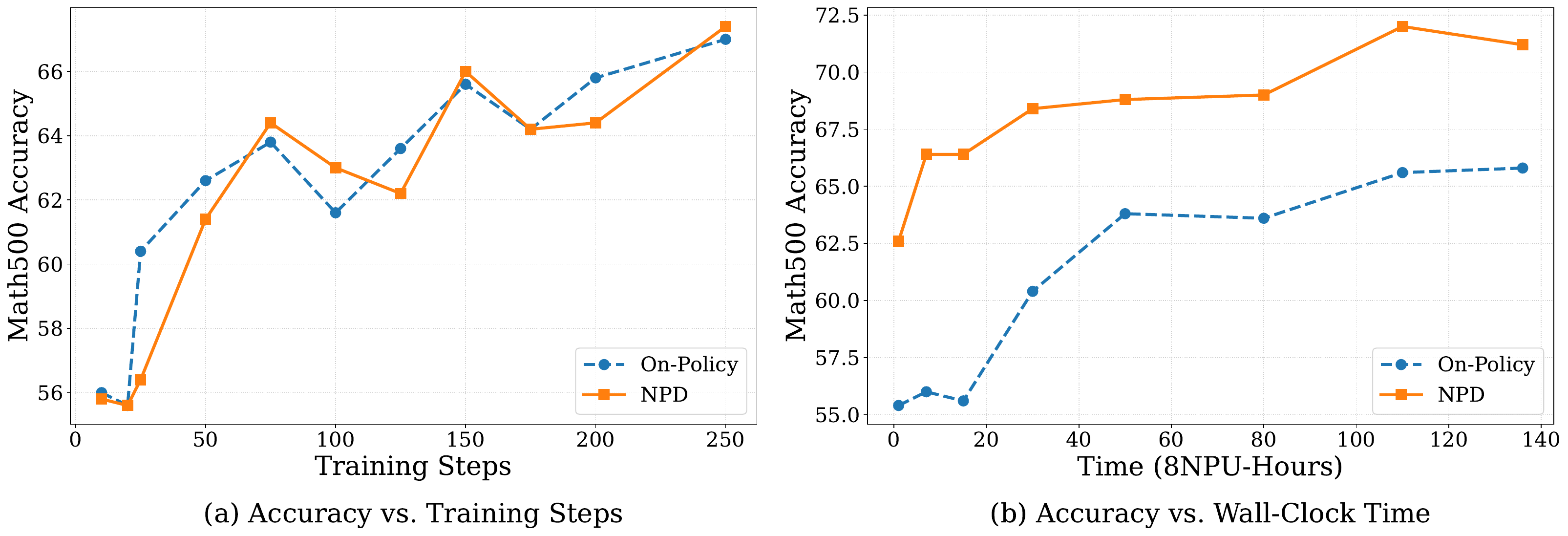}
    \caption{Comparison of NPD and On-Policy Distillation on Math500. \textbf{(a)} While both methods exhibit similar convergence trends, Near-Policy demonstrates superior data efficiency and maintains a consistent accuracy advantage. \textbf{(b)} Under fixed computational resources, Near-Policy achieves a substantial performance margin over On-Policy distillation.}
    \label{fig:efficiency_comparison}
    \vspace{-0.4cm}
\end{figure*}

\textbf{Performance under Fixed Steps.} 
Benefiting from SFT-compatible architecture, NPD enables sequence packing, which allows for significantly higher effective token throughput per step compared to the sample-wise padding often required by on-policy methods. As illustrated in the left of Fig~\ref{fig:efficiency_comparison}, under identical training steps, the accuracy trajectories on Math500 for both NPD and on-policy methods exhibit a highly consistent upward trend. This observation is critical: it demonstrates that NPD effectively retains the \textit{optimization quality} of on-policy methods. Despite the asynchronous nature of our data generation, the student's policy updates remain as precise and effective as those derived from synchronous, strictly on-policy gradients.

\textbf{Performance under Fixed Resources.} The practical advantage of NPD becomes even more pronounced when controlling for computational resources (i.e., identical wall-clock time). As shown in the right panel of Fig ~\ref{fig:efficiency_comparison}, NPD's performance curve consistently surpasses the on-policy baseline. By decoupling the generation phase and pre-computing Student-Generated Outputs, NPD eliminates the computational bottleneck associated with real-time, step-by-step sampling. Leveraging the inherent throughput efficiency of the SFT architecture, NPD processes significantly more data within the same time window, thereby establishing a substantial performance margin over on-policy baselines in real-world training scenarios.

% While strict on-policy algorithms (e.g., PPO) theoretically minimize distributional shift through continuous updates—often at prohibitive computational costs—our results demonstrate an asymptotic convergence behavior. The rapid plateau in performance gains suggests that the theoretical upper bound of on-policy optimization can be effectively approximated via sparse, periodic updates, thereby achieving a superior balance between efficacy and efficiency.

\begin{table}[htbp]
    \vspace{-0.6cm}
    \centering
    \caption{Ablation study of data filtering strategies within NPD. The best result is \textbf{bolded}.}
    \label{tab:ablation_data_filter}
    \small
    \setlength{\tabcolsep}{5pt} 
    
    \begin{tabular}{l c c}
        \toprule
        \textbf{Filtering Metric} & \textbf{Selection Scope} & \textbf{AVG} \\
        \midrule
        
        % 第一部分：基准与传统指标
        NPD (Unfiltered) & -- & 61.63 \\
        Perplexity (PPL) & Weighted & 61.69 \\ % 假设 weight 是指加权，或者改成具体的阈值
        $\text{IFD}_{\text{Teacher}}$ & $[0, 1]$ & 62.69 \\
        $\text{IFD}_{\text{Student}}$ & $[0, 1]$ & 62.60 \\

        % 使用 \cmidrule 替代 \midrule，并留一点空隙，表示逻辑分组但不割裂
        \cmidrule(lr){1-3} 
        
        % 第二部分：Delta 变体 (去掉 Ours 标题，直接写方法名)
        % 统一用 Delta 开头，显得整齐
        $\Delta_{\text{IFD}}$ (Single-side constraint) & $(0, \infty)$ & 61.77 \\
        $\Delta_{\text{IFD}}$ (Single-side constraint) & $(-\infty, 1)$ & 61.76 \\
        
        % 最终方案
        \textbf{$\Delta_{\text{IFD}}$ (Combined constraint)} & \boldmath$[0, 1]$ & \textbf{62.78} \\
        
        \bottomrule
    \end{tabular}
    \vspace{-0.4cm}
\end{table}

\subsection{Ablation on $\Delta$-IFD Filtering Strategy}

We conduct comprehensive ablation studies to evaluate the effectiveness of our data filtering mechanism. 

\textbf{Comparison of Filtering Metrics.} 
To validate the superiority of $\Delta_{\text{IFD}}$, we benchmark it against standard filtering strategies, including Perplexity (PPL)~\cite{jelinek1977perplexity}, Student-IFD, and Teacher-IFD. As detailed in Table \ref{tab:ablation_data_filter} , relying solely on absolute metrics yields sub-optimal results: PPL (61.69\%) tends to favor shorter, simpler sequences, while raw IFD scores ($\text{IFD}_{\text{Student}}$: 62.60\%, $\text{IFD}_{\text{Teacher}}$: 62.69\%) fail to capture the student's specific learning needs. In contrast, our proposed $\Delta_{\text{IFD}}$ achieves the highest baseline score of 62.78\%. This empirical evidence confirms that neither absolute confidence nor absolute difficulty is sufficient; capturing the \textit{relative knowledge gap} is critical for preventing policy lag and ensuring effective knowledge transfer in asynchronous distillation.

\begin{wrapfigure}{r}{0.4\textwidth} % r 表示靠右 (right)，0.5\textwidth 是占用一半宽度
    % \vspace{-0.2cm} 
    \centering
    \includegraphics[width=\linewidth]{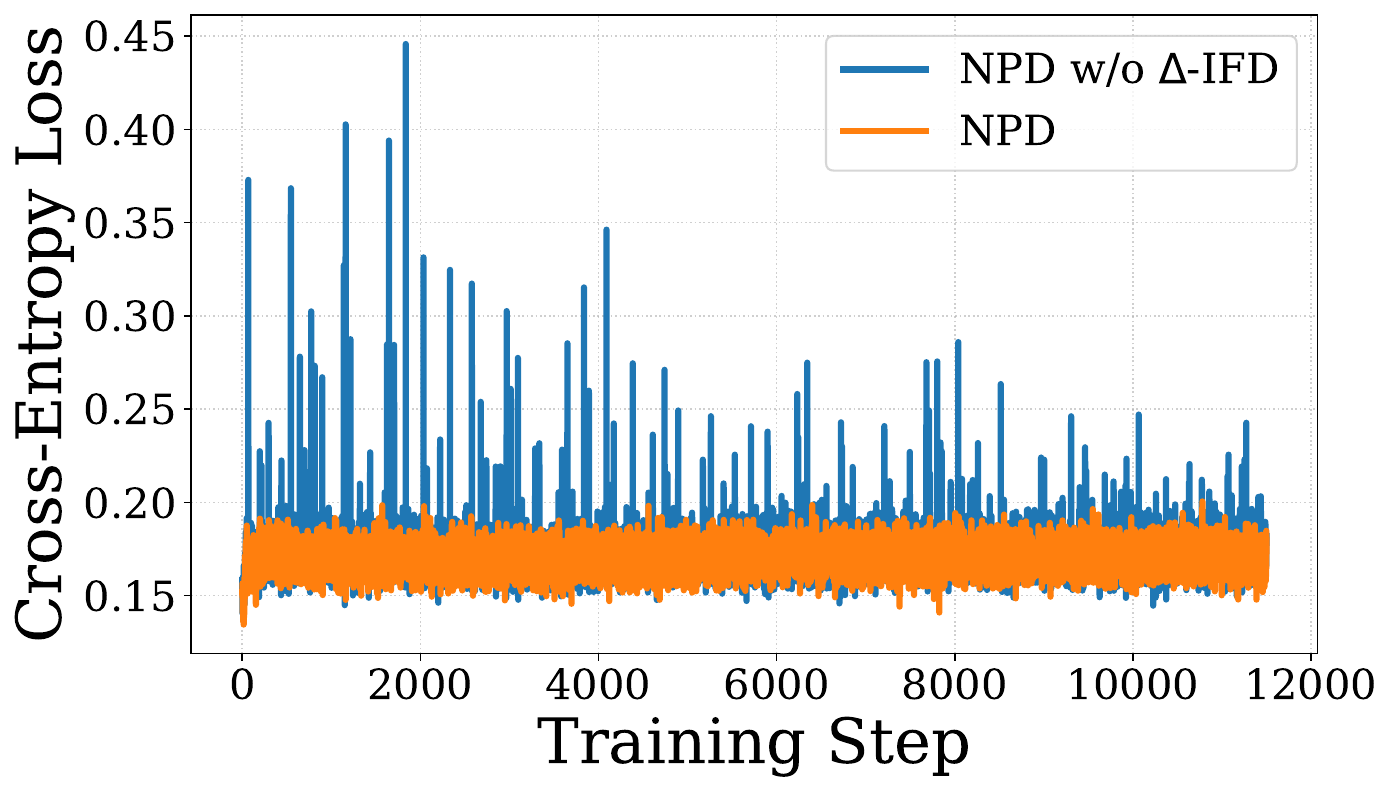}
    \caption{Comparison of training loss with and without $\Delta_{\text{IFD}}$ data filtering.}
    \label{fig:loss_compare}
    \vspace{-0.8cm} 
\end{wrapfigure}

\textbf{Constraint Analysis and Stability.} 
Building on the efficacy of $\Delta_{\text{IFD}}$, we further analyze its internal constraints. As shown in Table \ref{tab:ablation_data_filter}, we compare the performance impact of applying individual criteria ($\Delta_{\text{IFD}} < 0$ or $\Delta_{\text{IFD}} > 1$) against the combined constraint ($0 \le \Delta_{\text{IFD}} \le 1$). The results demonstrate that the combined strategy yields a 1.2\% improvement in accuracy over the baseline. This validates that removing both ``degenerate noise" ($\Delta_{\text{IFD}} < 0$) and ``cognitive overload" ($\Delta_{\text{IFD}} > 1$) is essential for optimal performance.

Furthermore, as illustrated in Fig~\ref{fig:loss_compare}, our filtering scheme effectively suppresses \textit{loss spikes} inherent to unfiltered training. By purging samples where the student is statistically overconfident or the teacher is uncertain, the $0 \le \Delta_{\text{IFD}} \le 1$ criterion significantly stabilizes the optimization landscape, leading to smoother convergence. Finally, sensitivity analysis (Table~\ref{tab:ablation_data_fliter_app}) further identifies $\tau=0.8$ as the optimal threshold for balancing data quantity and quality.

\begin{table}[htbp]
    \centering
    \vspace{-0.2cm}
    % ================= 左侧表格 (占据 55% 宽度) =================
    \begin{minipage}[t]{0.49\linewidth}
        \centering
        \caption{Performance and computational cost trade-off with different update frequencies. The best performance is highlighted in bold.}
        \label{tab:update_cost_tradeoff}
        \small
        \begin{tabular}{cccc}
            \toprule
            \textbf{Method} & \textbf{Update Freq.} & \textbf{AVG} & \textbf{NPU-Hours} \\
            \midrule
            \multirow{3}{*}{\textbf{NPD}} & 0 & 62.78 & 335 \\
                                          & 1 & \textbf{64.37} & 659 \\
                                          & 2 & 64.17 & 983 \\
            \bottomrule
        \end{tabular}
    \end{minipage}\hfill % \hfill 用于自动撑开中间的间距，使两侧对齐
    % ================= 右侧表格 (占据 40% 宽度) =================
    \begin{minipage}[t]{0.49\linewidth}
        \centering
        \caption{Ablation study of filtering strategies exploring threshold sensitivity, defined by the condition $0 \le \Delta_{\text{IFD}} \le \tau$. The best result is \textbf{bolded}.}
        \label{tab:ablation_data_fliter_app}
        \small
        \setlength{\tabcolsep}{6pt}
        \begin{tabular}{l c}
            \toprule
            \textbf{Filtering Strategy} & \textbf{AVG} \\
            \midrule
            $\tau = 0.9$ & 62.86 \\
            $\tau = 0.8$ & \textbf{63.05} \\
            $\tau = 0.7$ & 62.60 \\
            \bottomrule
        \end{tabular}
    \end{minipage}
    
    % 可选：统一调整整个并排模块与下方正文的间距
    % \vspace{-0.4cm} 
\end{table}

\subsection{The Efficacy of Sparse Updates}

Table~\ref{tab:update_cost_tradeoff} examines the critical trade-off between update frequency and computational overhead. Defining the update frequency as the number of policy refreshes per 10-epoch cycle, we observe a distinct pattern of \textit{diminishing marginal returns}. Specifically, a single strategic update (at epoch 5) yields a significant 1.59\% performance boost, whereas increasing the frequency to two or more leads to saturation or even slight regression. This phenomenon suggests that excessive synchronization not only incurs higher computational costs but may also introduce unnecessary perturbations that disrupt student learning stability.

These empirical findings strongly validate our \textbf{``Sparse Updates''} hypothesis. Contrary to strict on-policy paradigms (e.g., PPO) that demand frequent, high-cost updates to minimize distributional shift, our results demonstrate that the theoretical upper bound of on-policy optimization can be effectively approximated via \textit{sparse, periodic updates}. Consequently, we establish that frequent synchronization is not strictly necessary; instead, a strategic low-frequency schedule achieves a superior \textit{Pareto frontier}, reconciling high training efficacy with maximum system efficiency.

\begin{table*}[t]
    \centering
    \caption{\textbf{Performance comparison of RL fine-tuning outcomes.} We utilize the categorization style to group benchmarks by domain. ``AVG" denotes the macro-average across all 11 datasets. Gains in parentheses indicate improvement over the respective initialization baselines.}
    % \vspace{-0.2cm}
    \label{tab:rl_comparison_categorized}
    % \footnotesize
    
    % 使用 scriptsize 和紧凑间距以适应 13 列
    % \scriptsize
    \setlength{\tabcolsep}{5pt}
    \resizebox{\textwidth}{!}{%
    \begin{tabular}{l ccccc cc cc cc c}
        \toprule
        \multirow{2}{*}{\textbf{Method}} & \multicolumn{5}{c}{\textbf{General}} & \multicolumn{2}{c}{\textbf{Math}} & \multicolumn{2}{c}{\textbf{Reasoning}} & \multicolumn{2}{c}{\textbf{Code}} & \multirow{2}{*}{\textbf{AVG}} \\
        \cmidrule(lr){2-6} \cmidrule(lr){7-8} \cmidrule(lr){9-10} \cmidrule(lr){11-12}
        & \textbf{MMLU} & \textbf{CMMLU} & \textbf{C-Eval} & \textbf{IF-Eval} & \textbf{CLUEWSC} & \textbf{GSM8K} & \textbf{MATH} & \textbf{DROP} & \textbf{GPQA} & \textbf{MBPP} & \textbf{HumanEval} & \\
        \midrule
        
        % --- SFT Group ---
        \multicolumn{13}{l}{\textit{\textbf{SFT Initialization}}} \\
        \midrule
        SFT Baseline & 63.21 & 53.10 & \textbf{58.51} & \textbf{56.38} & 76.95 & 70.89 & 56.20 & 28.73 & 43.43 & 52.53 & \textbf{59.15} & 56.28 \\
        % + GRPO(200 steps) & \textbf{63.43} & 51.98 & 56.44 & \textbf{56.56} & 75.00 & \textbf{72.25} & \textbf{63.20} & \textbf{38.00} & 39.90 & \textbf{56.42} & 57.93 & 57.37 \tiny{(+1.1)} \\
        + GRPO (300 steps) & \textbf{63.43} & \textbf{53.65} & 57.15 & 54.16 & \textbf{77.05} & \textbf{77.26} & \textbf{72.00} & \textbf{45.64} & \textbf{43.94} & \textbf{56.81} & 57.32 & \textbf{59.86} \tiny{(+3.58)} \\
        \midrule
        
        % --- KD Group ---
        \multicolumn{13}{l}{\textit{\textbf{NPD Initialization (Ours)}}} \\
        \midrule
        NPD Baseline & 66.44 & 55.43 & 66.73 & \textbf{65.43} & 80.94 & 78.24 & 72.60 & 45.69 & \textbf{50.51} & 58.37 & 67.68 & 64.37 \\
        % + GRPO(200 steps) & \textbf{67.53} & 55.36 & 65.30 & 60.26 & 80.12 & \textbf{83.09} & \textbf{84.80} & \textbf{67.15} & \textbf{52.02} & \textbf{60.31} & \textbf{70.73} & \textbf{67.88} \tiny{(+3.5)} \\
        + GRPO (300 steps) & \textbf{68.04} & \textbf{56.53} & \textbf{67.13} & 60.81 & \textbf{82.17} & \textbf{87.60} & \textbf{84.76} & \textbf{69.18} & 48.48 & \textbf{61.87} & \textbf{69.51} & \textbf{68.73} \tiny{(+4.36)} \\
        \bottomrule
    \end{tabular}
    }
    % \vspace{-0.6cm}
\end{table*}

\begin{table*}[t]
    \centering
    \footnotesize
    \caption{Instruct model (non-thinking) comparison between \modelname{}-RL and other representative models across a diverse set of benchmarks for evaluating language and reasoning skills. \textbf{Bold} values represent the best results in each line among models at the 1B-parameter scale. If the original paper reports the results, we present the results from the original paper (marked with asterisks $^*$); otherwise, we list our reproduced results.}
    % \vspace{-0.2cm}
    \renewcommand{\arraystretch}{1.3}
    \resizebox{1.0\textwidth}{!}{
    \small
    \begin{tabular}{@{}c l | c c| c  c  c c| c c c}
    % \hline
    \noalign{\hrule height 1pt}
    & \multirow{2}{*}{\centering \textbf{Benchmark {\tiny (Metric)}}}   & \multirow{2}{*}{\centering\textcolor{black!100}{\textbf{Qwen3}}}   &\multirow{2}{*}{\centering\textcolor{black!100}{\textbf{Qwen2.5}}}   & \multirow{2}{*}{\centering\textbf{Gemma3}} & \multirow{2}{*}{\centering\textbf{Llama3.2}} & \multirow{2}{*}{\centering{\textbf{Qwen3}}} & \multirow{2}{*}{\centering\textbf{MiniCPM4}} & \multicolumn{3}{c}{\centering\textbf{openPangu-Embedded}} \\
    &   &   &   &   & &  &  & SFT & KD & RL \\
    \hline
    & \# Total Params  & \textcolor{black!100}{1.7B} & \textcolor{black!100}{1.5B} & 1B & 1B &0.6B & 0.5B &  1B &  1B & 1B \\
    \hline
    \multirow{5}{*}{\textbf{General}} & MMLU{\tiny (Acc)} &    \textcolor{black!100}{63.37} & \textcolor{black!100}{52.84}  & 37.49 & 32.19  & 44.24 & 55.55$^*$ &  63.21 &  66.44 & \textbf{68.04} \\
    & CMMLU {\tiny (Acc)}& \textcolor{black!100}{61.22}  & \textcolor{black!100}{53.98}   & 31.57 & 10.81 & 42.94 & \textbf{65.22}$^*$ &  53.10 & 55.43 & 56.53 \\
    
    & C-Eval {\tiny (Acc)}&  \textcolor{black!100}{61.00$^*$} & \textcolor{black!100}{59.30} & 32.49  & 32.08& 42.60$^*$  & 66.11$^*$ &  58.51 & 66.73 & \textbf{67.13} \\
    
    & IF-Eval {\tiny (Prompt Strict)} & \textcolor{black!100}{68.20$^*$} & \textcolor{black!100}{42.50$^*$} & 51.57  & 39.37 &54.50$^*$  & 50.28  & 56.38 & \textbf{65.43} & 60.81 \\
    
    & CLUEWSC{\tiny (Acc)}  & \textcolor{black!100}{77.36} & \textcolor{black!100}{74.59} &  50.20 & 52.36 & 50.31 & 49.90 &  76.95 & 80.94 & \textbf{82.87}\\
    
    \hline
    \multirow{2}{*}{\textbf{Math}} & GSM8K {\tiny (Acc)} & \textcolor{black!100}{77.03} & \textcolor{black!100}{73.20$^*$}  & 57.16 & 36.39 &  59.29 & 52.08$^*$ & 70.89 & 78.24 & \textbf{87.60} \\ 
    & MATH-500 {\tiny (Acc)} & \textcolor{black!100}{73.00$^*$} & \textcolor{black!100}{46.60}  & 39.40  & 18.20 & 55.20$^*$ & 29.60$^*$ & 56.20 & 72.60 &\textbf{84.76} \\
    \hline
    \multirow{2}{*}{\textbf{Reasoning}}& DROP {\tiny (F1)}  & \textcolor{black!100}{61.21} & \textcolor{black!100}{48.44} & 30.98  & 45.23 &  34.69 & 30.07 & 28.73 & 45.69 & \textbf{69.18}  \\ 
    & GPQA-Diamond {\tiny (Pass@1)}   & \textcolor{black!100}{28.60$^*$} & \textcolor{black!100}{29.80$^*$}  & 19.20$^*$ & 29.29 & 22.90$^*$ & 28.28 & 43.43 & \textbf{50.51} & 48.48 \\
    \hline
    \multirow{2}{*}{\textbf{Code}} & MBPP {\tiny (Pass@1)}   & \textcolor{black!100}{60.70} & \textcolor{black!100}{63.20$^*$}  & 58.75  & 40.08 &  46.69 & 59.14$^*$ & 52.53 & 58.37 & \textbf{61.87}  \\
    & HumanEval {\tiny (Pass@1)}  & \textcolor{black!100}{68.90} & \textcolor{black!100}{61.60$^*$} & 40.24  & 29.88  & 40.85 & 46.34$^*$ & 59.15 & 67.68& \textbf{69.51} \\
    \hline
    & Average & \textcolor{black!100}{63.69} & \textcolor{black!100}{55.10}   & 40.82 & 33.26 & 44.93 & 48.42 & 56.28  & 64.37 & \textbf{68.73} \\
    \noalign{\hrule height 1pt}
    \end{tabular}
    }
    \label{tab:benchmark-post-training}
    \vspace{-0.4cm}
\end{table*}

\subsection{Benefits for Subsequent RL}
To validate the hypothesis that NPD serves as a superior initialization for Reinforcement Learning, we conduct comparative experiments using Group Relative Policy Optimization (GRPO)~\cite{guo2025deepseek}. We initiate RL training from two distinct baselines: the standard SFT model and our NPD model, using a dataset of 7,000 mathematical problems.

As shown in Table~\ref{tab:rl_comparison_categorized}, under identical training budgets (300 steps), the NPD initialization yields significantly higher performance gains. While the SFT-based model achieves a modest improvement of 3.58\%, the NPD-based model achieves a substantial gain of 4.36\%. Specifically, on mathematical benchmarks, the NPD-initialized model boosts performance by 6.52\% on GSM8K and 15.00\% on Math500. Furthermore, we observe notable \textit{positive transfer} to out-of-domain tasks: despite the RL training focusing exclusively on math, the model exhibits improvements in general knowledge (MMLU +1.6\%) and code generation (MBPP +3.50\%, HumanEval +1.83\%).

\subsection{Compare with SOTA Models}
% We evaluate \modelname{}, which is trained using NPD and RL, on both reasoning and normal language tasks.

% Leveraging the computational efficiency and high fidelity of the Near-Policy strategy, we apply our 'NPD + RL' optimization framework to the openPangu-1B model. As detailed in Table ~\ref{tab:benchmark-post-training}, the model first establishes foundational capabilities during the SFT stage, achieving an average score of 56.28$\%$. Subsequently, the introduction of NPD effectively expands the solution space and optimizes the initial distribution, significantly elevating performance to 64.37$\%$. Finally, building upon this superior initialization, RL further unlocks the model's reasoning potential, propelling the final score to 68.78$\%$. This result not only validates the effectiveness of our method but also positions Pangu-1B to significantly outperform existing mainstream open-source models of comparable size, setting a new performance benchmark.

We evaluate \modelname{} on both reasoning and general language tasks by applying the combined NPD and RL framework to the openPangu-1B backbone. As detailed in Table~\ref{tab:benchmark-post-training}, the model exhibits a clear evolutionary trajectory: starting from an SFT baseline of 56.28\%, NPD optimizes the initial policy distribution to reach 64.37\%, and subsequent RL refinement further unlocks reasoning potential, propelling the final score to 68.73\%. This stepwise improvement validates that our framework effectively maximizes the potential of lightweight models, establishing a new performance benchmark for the 1B scale.

 A salient finding is its aggregate performance, achieving an average score of 68.73$\%$, significantly outperforming the larger Qwen3-1.7B model (63.69$\%$). This result highlights exceptional parameter efficiency, suggesting that advanced training and alignment methodologies can be more impactful than merely scaling model size. The model's advantage is most pronounced in mathematical and complex reasoning, where it secures leading scores on GSM8K (87.60$\%$) and MATH-500 (84.76$\%$). These strengths are complemented by robust general knowledge capabilities, as evidenced by top-tier performance on benchmarks such as CLUEWSC (82.87$\%$) and MMLU (68.04$\%$).

% \paragraph{Baselines \& Benchmarks.}  The \modelname~family is compared against several prominent open-source models within a similar parameter class to ensure a relevant and competitive analysis. The compared baselines include Qwen3 (1.7B and 0.6B)~\cite{yang2025qwen3}, Qwen2.5(1.5B)~\cite{Yang2024Qwen25TR}, Gemma3 (1B)~\cite{team2024gemma}, Llama3.2 (1B)~\cite{dubey2024llama} and  MiniCPM4 (0.5B)~\cite{team2025minicpm4}. Including the larger Qwen3-1.7B model serves as a critical point of comparison, allowing for an evaluation of the parameter efficiency of the proposed approach. Performance is measured using standard metrics appropriate for each benchmark. Accuracy (Acc) is used for tasks with single correct answers, such as MMLU~\cite{hendrycks2020measuring} and GSM8K~\cite{cobbe2021training}. The F1 Score is employed for tasks like DROP~\cite{Dua2019DROPAR}, which require a balance of precision and recall in text extraction. For code generation tasks like MBPP~\cite{austin2021program} and HumanEval~\cite{Chen2021EvaluatingLL}, Pass@1 is used, which measures the percentage of problems for which a correct solution is generated in a single attempt.

% \end{itemize}

% \paragraph{Evaluation Results.}

\vspace{-0.2cm}
\section{Conclusion and Discussion}

% We introduce NPD, a robust framework that effectively navigates the efficiency-quality trade-off by synergizing asynchronous generation with sparse student updates and $\Delta$-IFD filtering. By integrating NPD with RL, we achieve SOTA performance with a lightweight 1B-parameter model, \modelname{}-RL, which outperforms existing comparable scale baselines. By effectively reconciling high-performance AI with edge constraints, NPD offers a scalable path toward ubiquitous on-device intelligence and future edge optimizations.

We introduce NPD, a robust framework that effectively navigates the efficiency-quality trade-off by synergizing asynchronous generation with sparse student updates and $\Delta$-IFD filtering. By integrating NPD with RL, we achieve SOTA performance with a lightweight 1B-parameter model, \modelname{}-RL, which outperforms existing baselines of comparable scale. Reconciling high-performance AI with edge constraints, NPD offers a scalable path toward ubiquitous on-device intelligence and future edge optimizations.

\section{Limitation}
While Near-Policy Distillation (NPD) offers significant gains in efficiency and performance, the student model's efficacy remains inherently constrained by the teacher's distribution. Specifically, any biases or suboptimal logits provided by the teacher during complex reasoning steps may lead to the propagation of these systemic errors to the student.

\section{Impact Statement}
Near-Policy Distillation makes LLM alignment more compute-efficient by decoupling generation from training and enabling sequence packing, while retaining key benefits of on-policy distillation. This can lower the cost of training and deploying better-aligned smaller models. However, higher efficiency may also speed up scaling and deployment, so careful dataset governance and transparent filtering (e.g., $\Delta$-IFD) are important to reduce bias and misuse. We hope this work encourages more researchers to explore scalable near-policy training pipelines and principled data selection for efficient alignment.
%%%%%%%%%%%%%%%%%%%%%%%%%%%%%%%%%%%%%%%%%%%%%%%%%%%%%%%%%%%%
\newpage
\bibliographystyle{plainnat} % 或者使用 {unsrt} / {abbrvnat}，具体视模板要求而定，模板通常默认设好了样式
\bibliography{references}
\appendix

\newpage
\section{Algorithmic Details of Near-Policy Distillation}
The complete algorithmic flow of the Near-Policy Distillation (NPD) framework is detailed in Algorithm \ref{alg:iterative_kd}. Specifically, the pseudocode illustrates the cyclical process of asynchronous trajectory generation, rigorous sample filtering via $\Delta$-IFD, and the subsequent high-throughput optimization using sequence packing. This decoupled design ensures that both the student and teacher models operate at maximum hardware utilization while mitigating the policy lag inherent in asynchronous updates.
\begin{algorithm}
\caption{Iterative Near-Policy Distillation (NPD) with $\Delta$-IFD}
\label{alg:iterative_kd}
\small
\begin{algorithmic}[1]
\State \textbf{Input:} Teacher model $p_{\text{T}}$, Student model $p_{\text{S}}^\theta$, Prompt dataset $\mathcal{D}_x$, Pack length $L$
\State \textbf{Hyperparameters:} Learning rate $\eta$, Update interval $K$, Loss weights $\lambda$
\State \textbf{Initialize:} Student parameters $\theta_{S}$, Replay buffer $\mathcal{D}_{\text{train}} \leftarrow \emptyset$
\For{epoch $e = 1$ \textbf{to} $E$}
    
    \If{$(e-1) \pmod K = 0$} 
    \State \textit{\# Phase 1: Asynchronous Data Refresh}
        % \State Generate raw responses: $\mathcal{Y}_{\text{raw}} \sim p_{\text{S}}^\theta(\cdot \mid \mathcal{D}_x)$
        \State Generate student dataset: $\mathcal{D}_S \leftarrow \{ (x, y) \mid x \in \mathcal{D}_x, y \sim p_{\text{S}}^\theta(\cdot \mid x) \}$
        \State $\Delta$-IFD Filtering:$\mathcal{D}_{\text{select}} \leftarrow \{ (x,y) \in \mathcal{D}_S \mid \Delta_{\text{IFD}}(x,y) \text{ is valid} \}$
        \State Sequence Packing:$\mathcal{S}_{\text{pack}} \leftarrow \text{Pack}(\mathcal{D}_{\text{select}}, \text{Length}=L)$
        
        \State \textit{\# Phase 2: Teacher Annotation}
        \State Compute Teacher Logits (no grad): $\mathcal{O}_{\text{T}} \leftarrow p_{\text{T}}(\mathcal{S}_{\text{pack}})$
        \State Update Buffer: $\mathcal{D}_{\text{train}} \leftarrow \{ (S, O_{\text{T}}) \mid S \in \mathcal{S}_{\text{pack}}, O_{\text{T}} \in \mathcal{O}_{\text{T}} \}$
    \EndIf
    
    \State \textit{\# Phase 3: Student Optimization}
    \For{minibatch $(S, O_{\text{T}}) \in \mathcal{D}_{\text{train}}$}
        \State Compute Student Logits: $O_{\text{S}} \leftarrow p_{\text{S}}^\theta(S)$
        % \State $\mathcal{L}_{\text{KD}} \leftarrow \mathcal{D}_{\text{KL}}(\text{Softmax}(O_{\text{T}}) \parallel \text{Softmax}(O_{\text{S}}))$
        \State Compute Probs: $p_{\text{T}} \leftarrow \text{Softmax}(O_{\text{T}}), p_{\text{S}} \leftarrow \text{Softmax}(O_{\text{S}})$
        \State Compute KD Loss: $\mathcal{L}_{\text{KD}} \leftarrow \sum_{c \in \mathcal{V}_{\text{top-}k}} p_{\text{T}}(c) \log \frac{p_{\text{T}}(c)}{p_{\text{S}}(c)}$
        \State $\mathcal{L}_{\text{total}} \leftarrow (1-\lambda) \mathcal{L}_{\text{CE}}(S) + \lambda \mathcal{L}_{\text{KD}}$
        \State Update: $\theta_{S} \leftarrow \theta_{S} - \eta \nabla_\theta \mathcal{L}_{\text{total}}$
    \EndFor
\EndFor
\State \textbf{Return} Optimized $\theta_{S}$
\end{algorithmic}
\end{algorithm}

\section{Comprehensive Ablation Studies and Hyperparameter Analysis}
\label{sec:appendix_ablation}

\subsection{KD Method Ablation}
As shown in Table~\ref{tab:distillation_methold_detail}, we benchmark several sequence-packed distillation strategies under the same teacher--student setting (\textbf{openPangu-Embedded-7B} $\rightarrow$ \textbf{openPangu-Embedded-1B}) on a 1.2M-sample real-world industrial dataset, with all models trained for 10 epochs. The results indicate that vanilla SFT provides a solid baseline but yields limited improvements on harder capabilities such as math, reasoning, and code. Both supervised KD and SeqKD bring consistent gains over SFT, suggesting that richer teacher supervision and sequence-level packing can enhance the student's generalization across diverse tasks. Notably, our \textbf{NPD} method achieves the best overall performance (AVG=64.37) and delivers broad improvements across most benchmarks, demonstrating its effectiveness and robustness in large-scale industrial distillation.

\begin{table*}[htbp]
    \centering
    \caption{Comprehensive comparison between NPD and sequence-packed distillation methods. The experiments utilize \textbf{openPangu-Embedded-7B} as the teacher and \textbf{openPangu-Embedded-1B} as the student. Evaluations are conducted on a large-scale real-world industrial dataset comprising 1.2 million samples, with all models trained for 10 epochs.}
    \label{tab:distillation_methold_detail}
    \resizebox{\linewidth}{!}{%
    \begin{tabular}{lcccccccccccc}
        \toprule
        \multirow{3}{*}{\textbf{Method}} & \multicolumn{5}{c}{\textbf{General}}
        & \multicolumn{2}{c}{\textbf{Math}} 
        & \multicolumn{2}{c}{\textbf{Reasoning}}
        & \multicolumn{2}{c}{\textbf{Code}} 
        & \multirow{3}{*}{\textbf{AVG}} \\
        \cmidrule(lr){2-6} \cmidrule(lr){7-8} \cmidrule(lr){9-10} \cmidrule(lr){11-12}
         & \textbf{MMLU} & \textbf{CMMLU} & \textbf{C-Eval} & \textbf{IF-Eval} & \textbf{CLUEWSC} & \textbf{GSM8K} & \textbf{MATH-500} & \textbf{DROP} & \textbf{GPQA} & \textbf{MBPP} & \textbf{HumanEval} &  \\
        \midrule
        SFT & 63.21 & 53.10 & 58.51 & 56.38 & 76.95 & 70.89 & 56.20 & 28.73 & 43.43 & 52.53 & 59.15 & 56.28 \\ 
        Supervised KD & 62.68 & 53.97 & 58.83 & 59.70 & 77.56 & 72.10 & 65.20 & 30.24 & 46.46 & 57.20 & 62.80 & 58.79  \\ 
        SeqKD & 61.74 & 54.94 & 62.83 & 59.89 & 79.92 & 74.30 & 61.20 & 36.59 & 47.98 & 55.25 & 60.37 & 59.55  \\
        \textbf{NPD (Ours)} & \textbf{66.44} & \textbf{55.43} & \textbf{66.73} & \textbf{65.43} & \textbf{80.94} & \textbf{78.24} & \textbf{72.60} & \textbf{45.69} & \textbf{50.51} & \textbf{58.37} & \textbf{67.68} & \textbf{64.37} \\
        \bottomrule
    \end{tabular}%
    }
\end{table*}

\subsection{NPD Component Ablation}
We perform a step-by-step ablation to isolate the contributions of each component. As shown in Table~\ref{tab:ablation}, transitioning from standard KD to Packed Distillation based on student responses provides the initial structural baseline jump. However, our two novel components ($\Delta$-IFD and Sparse Updates) provide crucial, cumulative gains necessary to reach state-of-the-art performance.

While the shift to student-generated packing establishes the foundation, the \textbf{Sparse Update} mechanism contributes the most among our novel additions (+1.59), as it physically resets the policy lag and prevents asynchronous training collapse. The $\Delta$-IFD filter provides an additional significant boost (+1.15) by strictly pruning harmful variance.

\begin{table}[htbp]
    \centering
    \caption{Ablation study of different modules in NPD. This table illustrates the cumulative performance gains by progressively integrating key components: Packed Distillation, the $\Delta$-IFD filtering mechanism, and the Sparse Update strategy. The results demonstrate that while student-response-based packing establishes a robust baseline, our proposed $\Delta$-IFD and Sparse Update are crucial for reaching state-of-the-art performance.}
    \label{tab:ablation}
    \begin{tabular}{lcc}
        \toprule
        Method / Component & AVG & $\Delta$ Improvement \\
        \midrule
        Standard KD & 58.79 & -- \\
        Packed Distillation (Student Rollouts) & 61.63 & + 2.84 \\
        + $\Delta$-IFD Filter & 62.78 & + 1.15 \\
        + Sparse Update (\textbf{Full NPD}) & \textbf{64.37} & + 1.59 \\
        \bottomrule
    \end{tabular}
\end{table}

\subsection{Sensitivity to Update Frequency}
Our framework demonstrates high robustness and is largely insensitive to the specific update frequency. As shown in Table~\ref{tab:update_frequency}, both the full NPD and the variant without the $\Delta$-IFD filter (NPD w/o IFD) achieve optimal performance with just one to two sparse updates (64.37 and 63.43, respectively).

Theoretically, a policy update should be dynamically triggered whenever the KL divergence approaches our predefined $\epsilon$ boundary (e.g., 0.10). However, our extensive empirical evaluations reveal that uniformly allocating 1 or 2 updates across a standard 10-epoch training cycle naturally intercepts the critical points just before severe policy drift occurs. Therefore, in practical deployments, it is not necessary to incur the computational overhead of dynamic monitoring; statically allocating one to two updates consistently yields optimal returns.

\begin{table}[htbp]
    \centering
    \caption{Sensitivity analysis of the update frequency ($\tau$). Both configurations peak at 1 or 2 updates, demonstrating that minimal interventions are sufficient to prevent policy lag.}
    \label{tab:update_frequency}
    \begin{tabular}{ccc}
        \toprule
        \textbf{Update Frequency ($\tau$)} & \textbf{NPD (Full)} & \textbf{NPD w/o IFD} \\
        \midrule
        0 & 62.78 & 61.63 \\
        1 & \textbf{64.37} & 63.26 \\
        2 & 64.17 & \textbf{63.43} \\
        3 & -- & 63.12 \\
        \bottomrule
    \end{tabular}
\end{table}

\subsection{NPD Train Hyperparameter Ablation}
We conduct a systematic ablation study on knowledge distillation to further enhance model performance. Our analysis focuses on two key dimensions: (i) the weighting of the distillation loss term, and (ii) the effect of the top-$k$ value during decoding.

\paragraph{KD Loss Weight.}
To optimize the balance between the standard cross-entropy loss and the distillation loss, we conduct experiments with different values of the KD loss weight ($\lambda_{KD}$), ranging from 0.5 to 1.0. To accelerate the experimental cycle, we conduct distillation experiments based on a single stage (Direct Fast Response) of SFT. This approach introduces a new Stage~2 to perform distillation guided by labels, with the distillation learning rate kept consistent with that used in SFT. This systematic comparison shows setting $\lambda_{KD}=0.9$ yields the best overall performance, as shown in Table~\ref{tab:distillation_loss_weight}.

\begin{table}[htbp]
    \centering
    \caption{Effect of KD loss weight on the average benchmark performance. The evaluation metric is the average zero-shot accuracy across eight benchmarks. All models use greedy decoding with a decode length of 8K.}
    \label{tab:distillation_loss_weight}
    \begin{tabular}{lcccccc}
        \toprule
        $\lambda_{KD}$ & 0.5 & 0.6 & 0.7 & 0.8 & \textbf{0.9} & 1.0 \\
        \midrule
        Average  & 53.94 & 54.07 & 53.88 & 54.43 & \textbf{55.29} & 54.44 \\
        \bottomrule
    \end{tabular}
\end{table}

\paragraph{Top-$k$ Value Effect.}
To explore the impact of the number of top tokens used in the distillation process, we experiment with four values for the top-$k$ parameter: 5, 10, 15 and 20. The experimental setup is consistent with that described in the KD Loss Weight section. We observe that training time remains almost identical across different $k$ values. In terms of performance, increasing $k$ from 5 to 10 yields an improvement of 0.51\%, with top-$k$=10 achieving the highest average accuracy. However, further enlarging $k$ to 15 or 20 leads to a degradation in performance, as shown in Table~\ref{tab:topk_value}. Hence, we adopt $\lambda_{KD}=0.9$ and top-$k$=10 by default unless otherwise specified.

\begin{table}[htbp]
    \centering
    \caption{Effect of top-$k$ value on the average benchmark performance. All models use greedy decoding with a decode length of 8K, and the KD loss weight is fixed at $\lambda_{KD}=0.9$.}
    \label{tab:topk_value}
    \begin{tabular}{lcccc}
        \toprule
        Top-$k$ & 5 & \textbf{10} & 15 & 20 \\
        \midrule
        Average Performance & 54.78 & \textbf{55.29} & 54.24 & 54.40 \\
        \bottomrule
    \end{tabular}
\end{table}

% ==========================================
\section{Theoretical and Empirical Analysis}
\label{sec:appendix_theoretical_analysis}

\subsection{Analysis of Policy Latency}
\begin{figure}[htbp]
    \centering
    \includegraphics[width=0.5\linewidth]{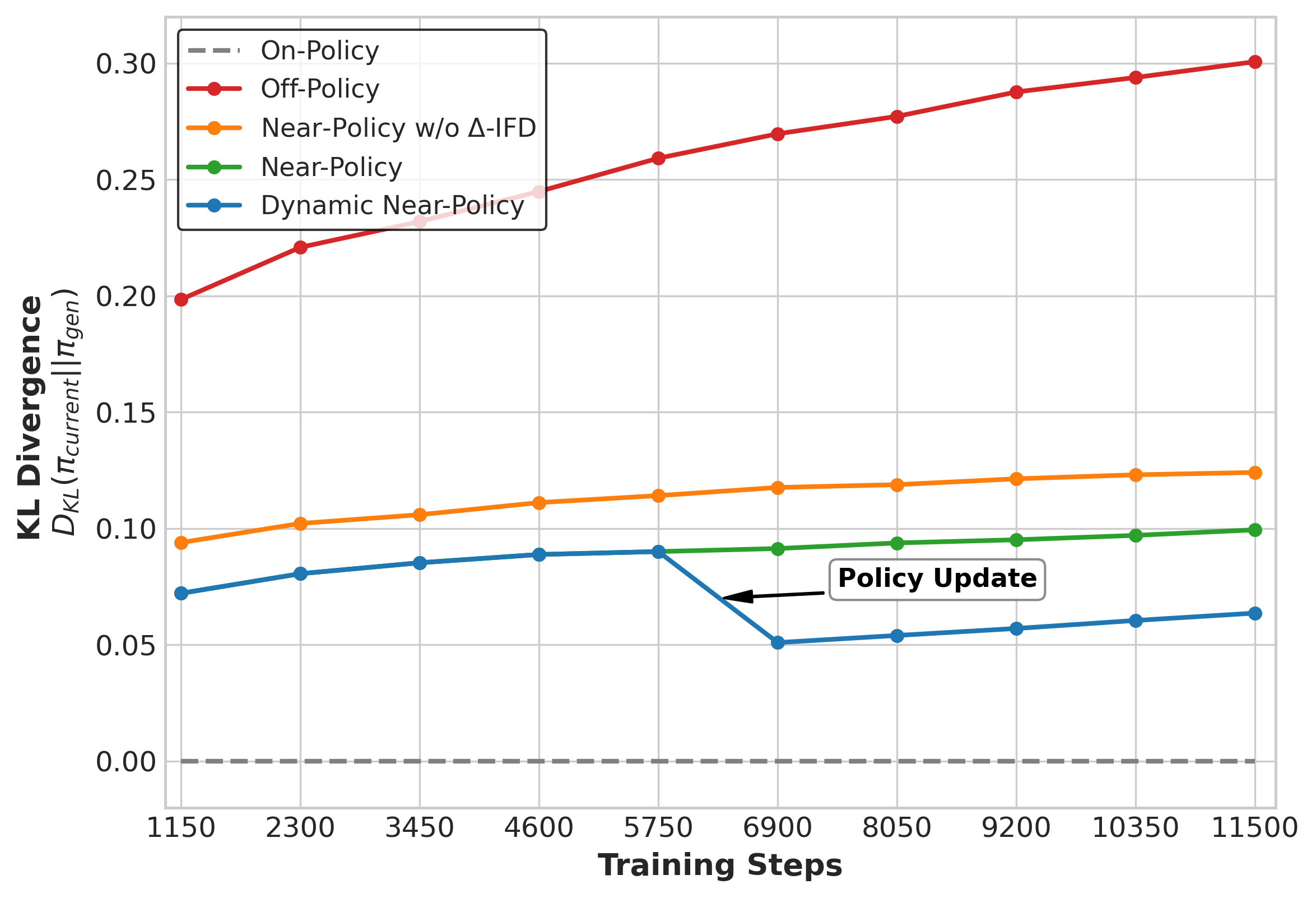}
    \caption{KL divergence over training steps. Unlike standard off-policy methods that suffer from severe drift, NPD maintains strict divergence bounds ($< 0.10$), with sparse updates ensuring forced policy synchronization.}
    \label{fig:policy_latency}
\end{figure}

To establish Near-Policy as a rigorous conceptual bridge, we conduct a quantitative analysis of policy latency. We track the Kullback-Leibler divergence, $D_{\text{KL}}(\pi_{\text{learner}} \parallel \pi_{\text{generator}})$, between the active learner and the rollout generator throughout the training process. As shown in Fig.~\ref{fig:policy_latency}, off-policy training exhibits severe distribution drift, with the KL divergence rapidly diverging to 0.30. In contrast, our proposed NPD method (green line) strictly maintains the KL divergence below 0.10. Furthermore, our sparse update mechanism (blue line) acts as a forced policy synchronization. When a weight sync occurs (e.g., around step 6900), the KL divergence sharply drops from 0.09 to 0.05, effectively resetting the policy lag.

Based on our empirical results, we formally define the \textbf{Near-Policy} regime: At any training step $t$, the distance between the current student policy ($\pi_{\text{current}}$) and the data-generating policy ($\pi_{\text{gen}}$) is strictly bounded by a small threshold $\epsilon$:
$$D_{KL}(\pi_{\text{current}}^{(t)} \parallel \pi_{\text{gen}}^{(t)}) \le \epsilon$$
Unlike strictly \textbf{on-policy} training ($\epsilon = 0$), which is computationally prohibitive, or \textbf{off-policy} training ($\epsilon \to \infty$), which leads to divergence, NPD dynamically enforces this boundary. Our results show that NPD empirically maintains a strict boundary of $\epsilon \approx 0.10$.

\subsection{Justification for $\Delta$-IFD}
Although $\Delta$-IFD may appear heuristic at first glance, it fundamentally serves as a measure of \emph{local policy alignment}. Since responses are actively sampled from the student's current policy, the student-side IFD is expected to satisfy a natural lower-bound property. When $\Delta$-IFD $< 0$ (i.e., Teacher IFD $<$ Student IFD), a theoretical \emph{statistical anomaly} arises: the student becomes paradoxically less confident in its own sampled response than the teacher. This indicates a mismatch between the sampled trajectory and the student's local policy geometry, suggesting that such samples may be unreliable for stable optimization.

As shown in Table~\ref{tab:ifd_filtering_ratio}, approximately 96\% of rollouts are safely retained, while only a small fraction of anomalous samples are filtered out. Importantly, this filtered 4\% corresponds to a high-variance and potentially poisonous long tail. Enforcing updates on these anomalous samples can lead to explosively compounding gradient variance and, in turn, destabilize training. Therefore, our design emphasizes \emph{precision over recall}: we prefer conservative filtering to reliably prevent catastrophic drift.

\begin{table}[htbp]
    \centering
    \caption{Filtering ratio of different rollout zones on the 12M proprietary dataset.}
    \label{tab:ifd_filtering_ratio}
    \begin{tabular}{lcc}
        \toprule
        \textbf{Zone} & \textbf{Early (\%)} & \textbf{Late (\%)} \\
        \midrule
        Degeneration         & 2.60  & 3.77 \\
        Proximal Learning    & 96.20 & 96.19 \\
        Cognitive Disconnect & 1.20  & 0.04 \\
        \bottomrule
    \end{tabular}
\end{table}

\subsection{Search Space Reduction}
To provide quantitative evidence for the search-space reduction effect, we analyze the autoregressive generation space of LLMs, which closely resembles the standard reinforcement learning setting where unconstrained optimization often leads to high-variance updates. In contrast, NPD actively prunes this space and restricts the optimizer to a high-reward sub-manifold.

\textbf{KL Divergence.} Our updated KL plot shows that unconstrained trajectories diverge rapidly. In comparison, $\Delta$-IFD imposes a strict bound on the optimization trajectory, empirically demonstrating that learning is confined to a narrow trust region.

\textbf{Causal Mechanism.} Removing $\Delta$-IFD significantly weakens the RL gains, reducing the improvement from $+4.36$ in full NPD to only $+2.48$ (see Table~\ref{tab:search_space_reduction}). This directly supports our claim that spatial pruning is a key mechanism for obtaining a higher-quality initialization and enabling more effective downstream RL optimization.

\begin{table}[htbp]
    \centering
    \caption{Effect of $\Delta$-IFD on initialization quality and downstream GRPO improvement.}
    \label{tab:search_space_reduction}
    \begin{tabular}{lc}
        \toprule
        \textbf{Method} & \textbf{AVG} \\
        \midrule
        NPD w/o $\Delta$-IFD Init & 63.43 \\
        + GRPO (300 steps) & 65.91 (+2.48) \\
        \midrule
        NPD Init (Ours) & 64.37 \\
        + GRPO (300 steps) & 68.73 (+4.36) \\
        \bottomrule
    \end{tabular}
\end{table}

% \begin{table}[htbp]
%     \centering
%     \caption{Comparison of SFT and NPD on $pass$@$k$ ($k$ = 1, 5, 10) on AIME24, AIME25, and CNMO2024.}
%     \label{tab:kd_rl_initialization}
%     \small
%     \setlength{\tabcolsep}{5pt} % 缩小列间距以适应单栏宽度
    
%     \begin{tabular}{l ccc ccc ccc}
%         \toprule
%         \multirow{2}{*}{\textbf{Method}} & \multicolumn{3}{c}{\textbf{AIME24}} & \multicolumn{3}{c}{\textbf{AIME25}} & \multicolumn{3}{c}{\textbf{CNMO2024}} \\
%         \cmidrule(lr){2-4} \cmidrule(lr){5-7} \cmidrule(lr){8-10}
%         & \textbf{@1} & \textbf{@5} & \textbf{@10} & \textbf{@1} & \textbf{@5} & \textbf{@10} & \textbf{@1} & \textbf{@5} & \textbf{@10} \\
%         \midrule
%         SFT & 14.31 & 27.59 & 27.59 & 11.37 & 13.33 & 16.67 & 12.30 & 16.67 & 16.67 \\
%         \textbf{NPD} & \textbf{19.58} & \textbf{27.59} & \textbf{34.48} & \textbf{20.00} & \textbf{26.67} & \textbf{26.67} & \textbf{16.17} & \textbf{33.33} & \textbf{44.44} \\
%         \bottomrule
%     \end{tabular}
%     % \vspace{-0.4cm} % 如果依然需要缩减下方留白，取消注释
% \end{table}

% ==========================================
\section{Extended Evaluations and Scalability}
\label{sec:appendix_extended_evaluations}

\subsection{Scalability to Larger Model Capacities}
To evaluate the scalability of our approach, we apply the NPD framework to larger-capacity models. Specifically, we distill a 72B teacher model into a 7B student model (\texttt{openPangu-7B}) using our internal 1.2M dataset. As shown in Table~\ref{tab:scaling_7b}, NPD yields significant improvements over the SFT baseline across multiple challenging benchmarks. This indicates that NPD scales effectively and maintains its performance advantages under larger parameter regimes.

\begin{table}[htbp]
    \centering
    \caption{Scalability experiments on a 7B student model. We compare the standard SFT baseline with our NPD framework when distilling from a 72B teacher.}
    \label{tab:scaling_7b}
    \begin{tabular}{lccccc}
        \toprule
        Method & AVG & GPQA & AIME24 & AIME25 & LiveCodeBench \\
        \midrule
        SFT & 38.32 & 63.64 & 25.21 & \textbf{35.00} & 29.41 \\
        \textbf{NPD (Ours)} & \textbf{41.73} & \textbf{69.36} & \textbf{32.08} & 33.12 & \textbf{32.35} \\
        \bottomrule
    \end{tabular}
\end{table}

\subsection{Thinking Models}
Beyond the fast-thinking base models used in our main study, we further evaluate NPD on a reasoning model setting using \textbf{openPangu-1B-think} as the student and \textbf{openPangu-7B-think} as the teacher, with the context window extended to 32K tokens.

As shown in Table~\ref{tab:thinking_models}, NPD delivers strong gains across diverse benchmarks and improves the overall average score from 66.43 to 73.52. Notably, even at a 32K context length, NPD still yields a large gain of \textbf{+7.09 AVG}, further validating its effectiveness for long-form reasoning models.

\begin{table*}[htbp]
    \centering
    \caption{Performance on a 1B student model with thinking mode with 32K context length.}
    \label{tab:thinking_models}
    \resizebox{\textwidth}{!}{
    \begin{tabular}{lcccccccccccc}
        \toprule
        \textbf{Method} & \textbf{MMLU} & \textbf{CMMLU} & \textbf{C-Eval} & \textbf{GSM8K} & \textbf{MATH-500} & \textbf{MBPP} & \textbf{HumanEval} & \textbf{GPQA} & \textbf{DROP} & \textbf{CLUEWSC} & \textbf{IFeval} & \textbf{AVG} \\
        \midrule
        SFT & 70.61 & 57.61 & 63.08 & 78.32 & 73.00 & 60.70 & 60.37 & 46.46 & 75.92 & 79.92 & 64.70 & 66.43 \\
        \textbf{NPD (Ours)} & \textbf{79.31} & \textbf{62.81} & \textbf{63.57} & \textbf{88.10} & \textbf{77.00} & \textbf{78.99} & \textbf{82.93} & \textbf{49.49} & \textbf{79.94} & \textbf{84.43} & \textbf{62.11} & \textbf{73.52} \\
        \bottomrule
    \end{tabular}
    }
\end{table*}

\subsection{Evaluation on Code Generation Tasks}
To further validate the efficacy of NPD, we conduct experiments in the coding domain using \texttt{Qwen2.5-Coder-7B-Inst} (Teacher) and \texttt{Qwen2.5-Coder-1.5B} (Student) on the \textit{WizardCoder} dataset. As summarized in Table~\ref{tab:coding_benchmarks}, NPD consistently outperforms existing baselines, demonstrating its robustness and generalization capabilities across diverse domains.

\begin{table}[htbp]
    \centering
    \caption{Performance comparison on coding benchmarks. NPD establishes a new state-of-the-art among baseline methods.}
    \label{tab:coding_benchmarks}
    \begin{tabular}{lccc}
        \toprule
        Method & HumanEval & MBPP & AVG \\
        \midrule
        $\mathcal{M}_T$ (Teacher) & 75.61 & 74.60 & 75.10 \\
        $\mathcal{M}_S$ (Student) & 30.73 & 60.84 & 45.78 \\
        \midrule
        GKD & 40.85 & 61.90 & 51.38 \\
        DistiLLM & 39.63 & 62.17 & 50.90 \\
        DistiLLM-2 & 42.24 & 62.70 & 52.47 \\
        \textbf{NPD (Ours)} & \textbf{43.58} & \textbf{63.01} & \textbf{53.29} \\
        \bottomrule
    \end{tabular}
\end{table}

\subsection{Multi-Path Reasoning Expansion}
To further push performance boundaries, we introduce an optional multi-path reasoning strategy. When computational budgets permit, we extend the student's inference process from generating a single greedy path to sampling $K$ independent trajectories for each instruction. This expansion significantly broadens sample diversity, allowing the model to explore a wider region of its potential solution space. Crucially, this approach synergizes effectively with our $\Delta$-IFD filtering mechanism to establish a robust ``\textit{generate-then-filter}'' paradigm: by rigorously applying the $\Delta_{\text{IFD}}$ criterion, we ensure that the distillation process benefits from a richer variety of reasoning patterns while strictly excluding low-quality noise.

Table~\ref{tab:strategy_ablation_checkmark} presents a stepwise ablation to decouple these contributions. Starting from the baseline (61.63\%), introducing Multi-Path Reasoning ($K=3$: 1 greedy, 2 sampled) boosts accuracy to 62.82\%. However, this generation overhead significantly increases training time. Subsequently, applying our $\Delta_{\text{IFD}}$ filter to these diverse trajectories further elevates performance to 63.29\%. This validates that while increasing sample diversity is beneficial, rigorously filtering out low-quality noise is essential for stabilizing optimization.

\begin{table}[htbp]
    \centering
    \caption{\textbf{Ablation study on the effectiveness of Multi-Path Reasoning and $\Delta_{\text{IFD}}$ Filtering.} We progressively incorporate the multi-path generation strategy and the dynamic filtering mechanism. \textbf{Train Time} indicates the computational overhead in NPU hours.}
    \label{tab:strategy_ablation_checkmark}
    \begin{tabular}{cccc} 
        \toprule
        \textbf{Multi-Path} & \textbf{$\Delta_{\text{IFD}}$} &  \textbf{Train Time (h)} & \textbf{AVG (\%)} \\
        \midrule
         &  & 335 & 61.63 \\
        \checkmark &  & 1000 & 62.82 \\
        \checkmark & \checkmark & 998 & \textbf{63.29} \\
        \bottomrule
    \end{tabular}
\end{table}

% ==========================================
\section{End-to-End Training Efficiency and Cost Analysis}
\label{sec:appendix_efficiency}

To provide full transparency regarding computational costs, we profile the end-to-end time breakdown for training NPD over 10 epochs on the 1.2M dataset (using an 8-NPU node):

\begin{table}[htbp]
    \centering
    \caption{End-to-end time breakdown for training NPD.}
    \label{tab:time_breakdown}
    \begin{tabular}{lcc}
        \toprule
        Phase & Time (8-NPU Node, h) & Percentage \\
        \midrule
        Student Rollout & 300 & 82.01\% \\
        Teacher Annotation & 18 & 4.92\% \\
        $\Delta$-IFD Computation & 5 & 1.37\% \\
        Filtering & 0.04 & 0.01\% \\
        Packing & 0.75 & 0.21\% \\
        Training & 42 & 11.48\% \\
        \bottomrule
    \end{tabular}
\end{table}

As shown in Table~\ref{tab:time_breakdown}, our $\Delta$-IFD filtering introduces negligible overhead (1.38\%), while Student Rollout dominates the cost (82.01\%). Traditional strict on-policy methods necessitate synchronous inference and optimization (i.e., a ``one-rollout-per-update'' paradigm), which severely underutilizes expensive training NPUs by leaving them in prolonged idle states. In contrast, NPD not only fully decouples the student rollout process asynchronously---enabling horizontal scaling with additional resources to accelerate inference---but also implements a ``single-rollout-multiple-updates'' strategy for data utilization. This architectural design substantially conserves overall computational resources and maximizes the system's training throughput.

%%%%%%%%%%%%%%%%%%%%%%%%%%%%%%%%%%%%%%%%%%%%%%%%%%%%%%%%%%%%

\newpage

\end{document}